\documentclass[runningheads,twocolumn]{llncs}

\usepackage{eccv}
\usepackage{eccvabbrv}

\usepackage{graphicx}
\usepackage{booktabs}
\usepackage{enumitem}
\usepackage{makecell}
\usepackage[table]{xcolor}
\usepackage[accsupp]{axessibility}
\usepackage{adjustbox}
\usepackage{microtype}
\usepackage[
  a4paper,
  hmargin=2.1cm,
  vmargin=2.65cm
]{geometry}

\makeatletter
\AtBeginDocument{%
  \setlength{\columnsep}{0.85cm}
}
\makeatother

\usepackage[breaklinks,colorlinks,citecolor=eccvblue]{hyperref}
\DeclareUnicodeCharacter{2212}{-}

\makeatletter
\renewenvironment{abstract}{%
  \list{}{\advance\topsep by0.5cm\relax\small
    \leftmargin=0cm
    \labelwidth=\z@
    \listparindent=\z@
    \itemindent\listparindent
    \rightmargin\leftmargin}%
  \item[\hskip\labelsep\bfseries\abstractname]}%
{\endlist}
\makeatother

\begin{document}

\title{FMS\texorpdfstring{\textsuperscript{2}}{2}: Unified Flow Matching for Segmentation and Synthesis of Thin Structures}

\titlerunning{FMS\textsuperscript{2}: Thin-Structure Learning}

\author{
{\Large Babak Asadi},
{\Large Peiyang Wu},
{\Large Mani Golparvar-Fard},
{\Large Viraj Shah},
{\Large Ramez Hajj}
}

\authorrunning{B. Asadi et al.}

\institute{
{\large University of Illinois at Urbana-Champaign}\\[6pt]
{\normalsize \email{\{basadi2,peiyang7,mgolpar,vjshah3,rhajj\}@illinois.edu}}\\[16pt]
}

\maketitle

\begin{abstract}

Segmenting thin structures like infrastructure cracks and anatomical vessels is a task hampered by topology-sensitive geometry, high annotation costs, and poor generalization across domains. Existing methods address these challenges in isolation. We propose FMS$^2$, a flow-matching framework with two modules. (1) \textbf{\emph{SegFlow}} is a 2.96M-parameter segmentation model built on a standard encoder-decoder backbone that recasts prediction as continuous image $\rightarrow$ mask transport. It learns a time-indexed velocity field with a flow-matching regression loss and outputs the mask via ODE integration, rather than supervising only end-state logits. This trajectory-level supervision improves thin-structure continuity and sharpness, compared with tuned topology-aware loss baselines, without auxiliary topology heads, post-processing, or multi-term loss engineering. (2) \textbf{\emph{SynFlow}} is a mask-conditioned mask $\rightarrow$ image generator that produces pixel-aligned synthetic image-mask pairs. It injects mask geometry at multiple scales and emphasizes boundary bands via edge-aware gating, while a controllable mask generator expands sparsity, width, and branching regimes. On five crack and vessel benchmarks, SegFlow alone outperforms strong CNN, Transformer, Mamba, and generative baselines, improving the volumetric metric (mean IoU) from 0.511 to 0.599 (+17.2\%) and reducing the topological metric (Betti matching error) from 82.145 to 51.524 (-37.3\%). When training with limited labels, augmenting SegFlow with SynFlow-generated pairs recovers near-full performance using 25\% of real annotations and improves cross-domain IoU by 0.11 on average. Unlike classical data augmentation that promotes invariance via label-preserving transforms, SynFlow provides pixel-aligned paired supervision with controllable structural shifts (e.g., sparsity, width, branching), which is particularly effective under domain shift. The code and a new large-scale dataset of thin-structure image--mask pairs (10k cracks, 1k vessels) will be released after acceptance.

\end{abstract}

\section{Introduction}
\label{sec:intro}

Segmenting thin structures is critical for a wide range of applications such as infrastructure inspection and medical imaging.
However, current methods still fall short of expected performance levels. 
Depending on image resolution, targets can be as thin as one pixel with surface sampling distance of $0.1-0.2$ \textit{mm/pixel}. These structures often occupy less than $1\%$ of the Field-of-View and the segmentation result must remain connected for downstream metrology and planning~\cite{mou2021cs2net,shit2021cldice,yang2023dconnnet,gupta2023topologyuncertainty,zuo2024crackmamba}. Pixel-wise annotation for ground truth is labor-intensive and exhibits inter-annotator variability, especially along low-contrast, branching structures and ambiguous margins~\cite{lesage2009review,zhao2021retvesselreview,rajitha2023retinalreview,zhou2023crackreview,wu2025crackreview}. Models trained on one dataset frequently fail to generalize when background texture, material, illumination, or acquisition protocol changes~\cite{isensee2018nnunet,chen2024transunet,wu2024medsegdiffv2,wu2025denver,yang2023dconnnet}. Taken together, (i) topological fragility at one-pixel scales, (ii) costly and variable supervision, and (iii) pronounced domain shift form the core challenges that most prior research tackle only in part, with most studies focusing on a single facet.

Architectural advances have improved in-domain accuracy. Multi-scale CNNs with explicit edge or contour heads~\cite{mou2021cs2net,acebes2024clce,xu2023dcsau,ouyang2025hybridvessel}, CNN–Transformer hybrids~\cite{chen2024transunet,zhou2025nnwnet} that fuse local detail with long-range context, and recent Mamba or SSM variants~\cite{liu2025scsegamba,zuo2024crackmamba,liu2024mambahrnet,song2025mambafuse} that propagate information at linear cost all contribute~\cite{cheng2022mask2former,qi2024unigs,wu2024medsegdiffv2}. Nevertheless, architecture alone is not a consistent remedy. On common thin-structure benchmarks, a vanilla U-Net baseline often matches~\cite{isensee2018nnunet,mou2021cs2net,zhou2023crackreview} and occasionally exceeds the performance of more elaborate architectures~\cite{zhou2025nnwnet,liu2025scsegamba,zuo2024crackmamba}. Two factors recur; the first is that limited dataset scale and severe class imbalance, which constrain the benefits of high-capacity attention or SSM modules and increase overfitting risk. In addition, overlap-style objectives do not enforce connectivity, so added capacity often models background texture rather than preserving continuous centerlines. Under cross-dataset evaluation, per-pixel classifiers trained with overlap objectives still show large performance drops~\cite{yang2023dconnnet,gupta2023topologyuncertainty,wu2025denver,zhou2023crackreview}.

\begin{figure}[!htbp]
  \centering
  \includegraphics[width=1\linewidth]{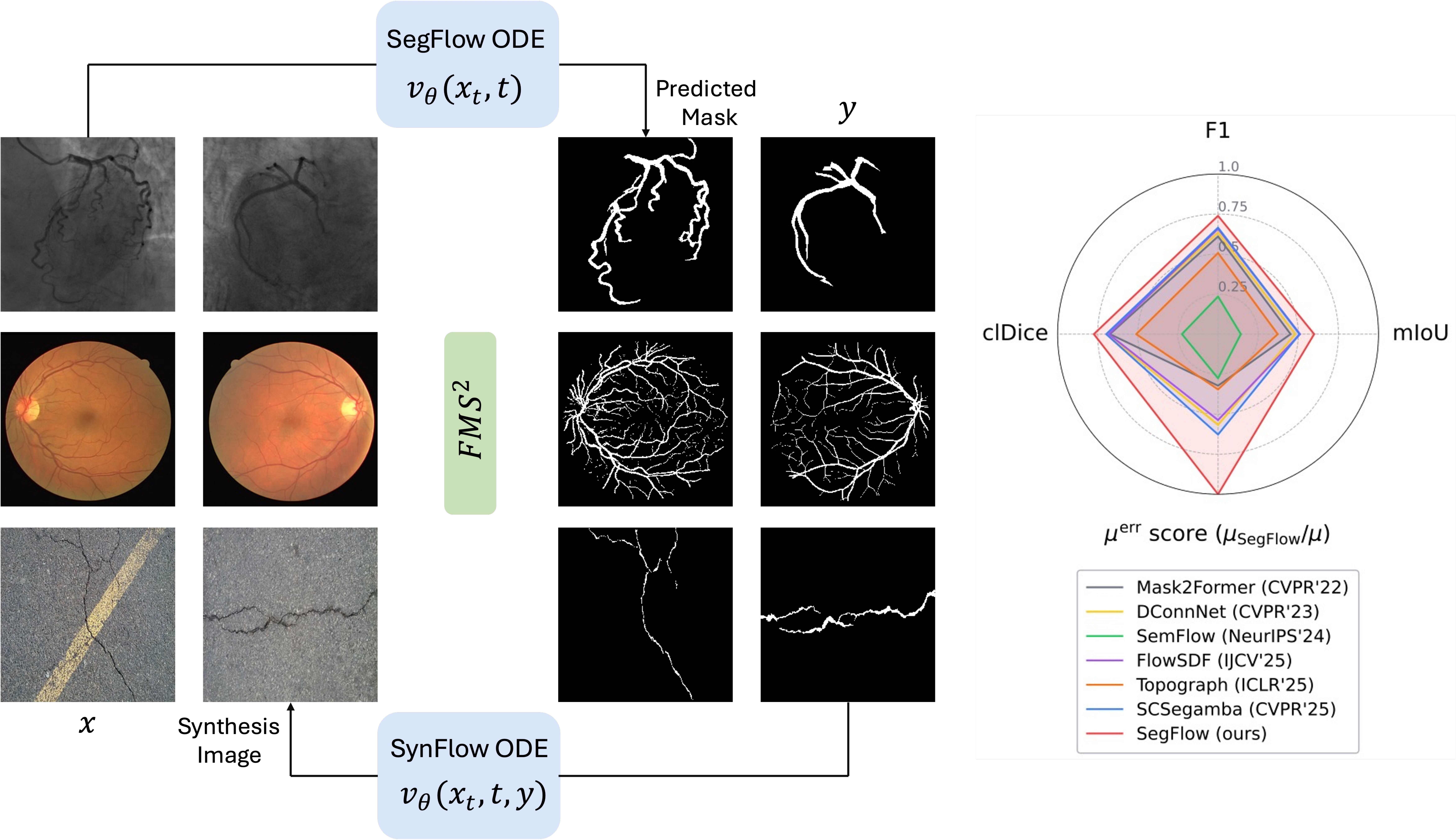} 
  \caption{\textbf{FMS$^2$ Framework.} SegFlow recasts segmentation as image$\rightarrow$mask transport problem,  effectively preserving sub-pixel topologies and yielding sharper segmentations than existing approaches. SynFlow is mask$\rightarrow$image renderer that generates topology-consistent image-mask pairs to alleviate the scarcity of labeled data for SegFlow training.\vspace{-3ex}}
  \label{fig:teaser}
\end{figure}

\vspace{0.7em}
A complementary thread introduces topology-aware supervision through differentiable losses such as clDice or through graph-based post-processing that reconnects broken centerlines~\cite{shit2021cldice,kirchhoff2024skeletonrecall,acebes2024clce,dong2025codice,lux2025topograph}. These approaches improve connectivity, but continue to supervise end-state logits rather than the formation process that yields continuous centerlines. 
These approaches also introduce additional complexity or memory requirements.

A third line reduces label cost or mitigates domain shift using synthetic data from GAN or diffusion models, or through frequency and geometry alignment with unlabeled targets~\cite{rombach2022ldm,wang2024semflow,qi2024unigs,wu2024medsegdiffv2,vangansbeke2024ldmseg,wu2025denver}. Generic conditioning pipelines can erase thin structures through normalization, which leads to mask–image mismatch, and iterative diffusion adds sampling overhead. When conditioning on provided masks, topological diversity can also be constrained by the mask distribution~\cite{zhang2023controlnet,yang2024scpcontrolnet,feng2025controlnetxs,yang2025crackdiffusion}.

A single framework that unifies accurate thin-structure segmentation, label efficiency, and robustness to domain shift is desirable. To this end, we introduce $\mathrm{FMS}^2$, a framework built on a \emph{flow-matching} formulation~\cite{lipman2022flowmatching,liu2023rectifiedflow,schusterbauer2025diff2flow,chen2024flowmatchingtutorial,zhang2024ictm} that systematically address these challenges. 

Our first contribution, \emph{SegFlow}, directly confronts the core problem of topological fragility (i). SegFlow recasts segmentation as an image$\rightarrow$mask transport problem. Instead of predicting per-pixel logits at a single end time, it learns a time-indexed velocity field that transports an input image toward its ground-truth mask under a deterministic flow. This approach supervises the \emph{entire formation process} of the mask, encoding topological consistency from the start of the trajectory rather than only at its final state. As our experiments show, SegFlow alone brings significant improvements in segmentation quality over strong baselines.

To address the second challenge of costly supervision (ii), we propose a complementary module, \emph{SynFlow}. This is a mask$\rightarrow$image renderer designed to enhance the label efficiency of the SegFlow model through synthetic paired data generation. SynFlow use the same U-Net architecture, but conditions on binary masks through multi-scale spatial modulation and an edge-aware gating path. This yields pixel-aligned synthetic image-mask pairs that preserve sub-pixel boundaries and counter the normalization-induced drift seen in other generative models~\cite{rombach2022ldm,zhang2023controlnet}. 
We also introduce a controllable mask generator that produces a diverse range of masks (from ultra-sparse to dense) to be fed into SynFlow, allowing us to concentrate synthetic supervision exactly where manual annotation is most expensive and topologically complex.

Training SegFlow with synthetic pairs from SynFlow lets the segmentation model inherit topology-aware transport, while the renderer supplies diverse, topology-consistent pairs. Together, our framework confronts (i) sub-pixel topology, (ii) limited supervision, and (iii) domain shift.

On five crack and vessel benchmarks, SegFlow alone delivers substantial improvements in both volumetric and topological measures over strong CNN, Transformer, Mamba, diffusion, and topology-aware baselines. It boosts the mean IoU from $0.511$ to $0.599$ (a \textbf{$17.2\%$} relative gain), raises the mean clDice from $0.697$ to $0.774$ (a \textbf{$11.05\%$} relative gain), and reduces the Betti matching error from 82.15 to 51.52 (−37.28\%). The addition of SynFlow provides significant gains in label efficiency and robustness. With only \textbf{$25\%$} human-annotated masks, training SegFlow with SynFlow-generated pairs recovers performance close to full supervision. Furthermore, under source-only cross-dataset evaluation, augmenting the source domain with SynFlow images improves target-domain IoU by about \textbf{0.11} on average. In summary, our primary contributions include:

\begin{itemize}[leftmargin=*,noitemsep,topsep=0pt,parsep=0pt,partopsep=0pt]
\item \textbf{\emph{SegFlow}}, a flow-matching segmentation model that substantially outperforms prior SOTA architectures and topology-aware loss designs for thin-structure segmentation, improving both volumetric overlap and topological fidelity without auxiliary heads or post-processing.

\item \textbf{\emph{SynFlow}}, a mask-conditioned renderer with multi-scale topology injection and an edge-aware-gating to produce pixel-aligned, topology-preserving synthetic image--mask pairs. Mixing small fraction of human-annotated data with SynFlow-generated data, substantially reducing label demand and improving cross-dataset robustness.

\item \textbf{A new large-scale dataset} of \textbf{10k crack} and \textbf{1k vessel} image--mask pairs to support benchmarking and the development of generalizable thin-structure models.
\end{itemize}

\begin{figure*}[t]
    \centering
    \includegraphics[width=1\textwidth]{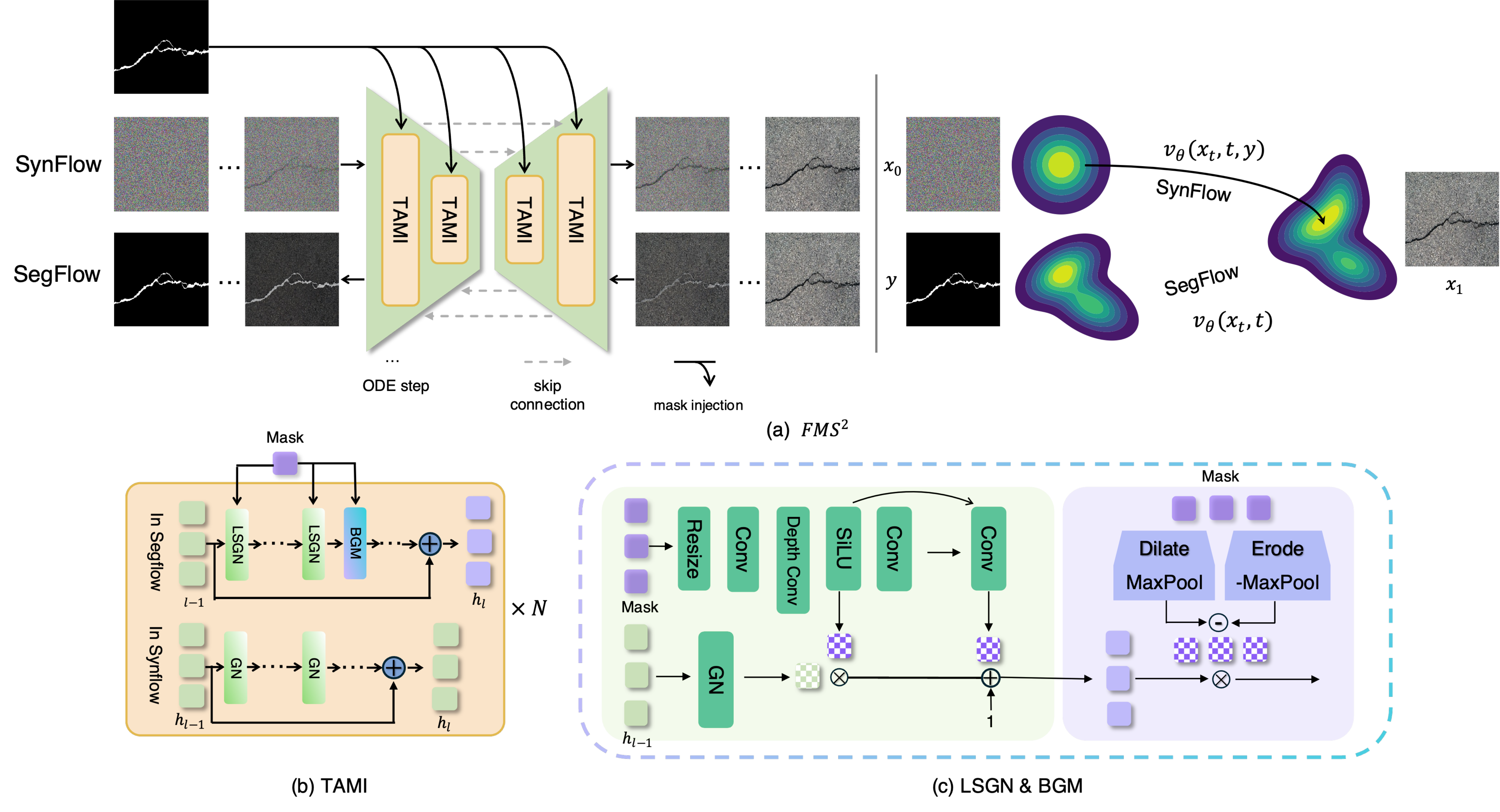}
    \caption{
\textbf{Architecture and transport view of FMS$^2$.}
\textbf{(a)} Coupled SegFlow (image$\rightarrow$mask) and SynFlow (mask$\rightarrow$image) share a U-Net–like backbone and are trained with flow matching: a time-indexed velocity field $v_\theta(x_t,t,\cdot)$ transports noise to images and masks, while multi-scale mask injection steers trajectories to preserve one-pixel topology.
\textbf{(b)} Toplogy-Aware Mask Injection (TAMI) used in both flows: in SynFlow it combines lite-SPADE GroupNorm (LSGN) with the boundary-gating module (BGM) to inject mask geometry and focus capacity near thin structures, while in SegFlow it reduces to GN-based modulation.
\textbf{(c)} LSGN uses a mask pyramid to generate affine parameters that modulate group-normalized features, whereas BGM builds an edge band via morphological dilate/erode and gates features near crack/vessel interfaces, emphasizing topology-critical regions.
}

    \label{fig:overview}
\end{figure*}

\section{Related Work}
\label{sec:related}

\subsection{Thin-structure segmentation}
Segmentation of thin structures spans generic encoder--decoders and topology-aware models. U\mbox{-}Net and nnU\mbox{-}Net remain strong baselines in biomedical and infrastructure domains~\cite{ronneberger2015unet,isensee2018nnunet}, and for cracks and vessels, multi-scale CNNs add edge or contour heads to stabilize one-pixel traces~\cite{mou2021cs2net,xu2023dcsau}. Transformer-based Mask2Former~\cite{cheng2022mask2former}, hybrid CNN--Transformer designs such as nnWNet~\cite{zhou2025nnwnet}, and vision SSMs like SCSegamba with structure-aware Mamba blocks~\cite{liu2025scsegamba} improve long-range context and efficiency, while DConnNet injects directional connectivity priors in the latent space for anatomy~\cite{yang2023dconnnet}. Yet cross-dataset studies still report sizable drops under appearance shift, suggesting that architecture alone does not guarantee topology preservation.

Topology-aware supervision makes connectivity explicit. clDice and its variants promote centerline overlap~\cite{shit2021cldice,acebes2024clce,dong2025codice}, Skeleton Recall Loss targets tubular connectivity with lower overhead~\cite{kirchhoff2024skeletonrecall}, Topology-Aware Uncertainty estimates structure-wise uncertainty~\cite{gupta2023topologyuncertainty}, and Topograph formulates a component-graph loss with preservation guarantees and efficient computation~\cite{lux2025topograph}. These approaches improve connectivity but still supervise only end-state logits, typically require dense labels plus extra loss terms or post-processing, and do not model the full evolution from image to mask.

\subsection{Generative models for segmentation \& synthesis}
Generative models support segmentation, label efficiency, and robustness. In medical imaging, MedSegDiff-V2 shows that diffusion-based segmentation models can outperform discriminative baselines in low-data regimes~\cite{wu2024medsegdiffv2}. Flow-based segmentation (e.g., FlowSDF) replaces stochastic denoising with deterministic flows tailored for segmentation. Flow-based segmentation has also appeared in FlowSDF~\cite{flowsdf}, which parameterizes supervision with an implicit SDF target and predicts it through noise-driven denoising. In contrast, SegFlow learns explicit pixel-space image$\rightarrow$mask transport with deterministic flow matching, providing a distinct supervision signal that better preserves sub-pixel branches. Joint segmentation--generation frameworks such as SemFlow~\cite{wang2024semflow} are conceptually related, but they learn a latent-space rectified flow that couples geometry and appearance, whereas we decouple the two and use SynFlow as a mask$\rightarrow$image renderer to generate \emph{paired, pixel-aligned} mask-conditioned samples with explicit multi-scale topology injection, which is particularly important for ultra-thin structures. Diffusion models enable high-fidelity synthesis~\cite{rombach2022ldm}, and latent-diffusion methods have been extended to panoptic label maps and mask inpainting~\cite{vangansbeke2024ldmseg}. Conditioning further improves controllability. ControlNet augments frozen diffusion with spatial conditions~\cite{zhang2023controlnet}, and later variants refine spatially adaptive normalization~\cite{yang2024scpcontrolnet,feng2025controlnetxs}. Latent-diffusion pipelines such as ControlNet~\cite{zhang2023controlnet} and T2I-Adapter~\cite{mou2024t2iadapter} typically encode masks or layouts via concatenation or auxiliary branches inside the U\mbox{-}Net latent path; recent work on curvilinear object segmentation reports that this latent conditioning can gradually wash out thin structures and blur topology~\cite{lei2024enriching}. 

Flow matching and rectified flows learn time-indexed velocity fields for deterministic transport~\cite{lipman2022flowmatching,liu2023rectifiedflow}. SemFlow binds segmentation and image synthesis via a rectified-flow ODE~\cite{wang2024semflow}, and Diff2Flow transfers diffusion quality to flows for faster sampling~\cite{schusterbauer2025diff2flow}. However, these diffusion/flow pipelines and their latent-conditioning schemes are tuned for semantic regions or thicker structures and remain sub-optimal for geometry-sensitive cracks and vessels, as they do not explicitly encode ultra-sparse, sub-pixel topology or mask-conditioned synthesis under controllable sparsity.

\section{Methodology}
\label{sec:met}

\subsection{Preliminaries}
\label{sec:prelim}
Flow Matching (FM) learns a time–indexed vector field that moves a state along a continuous path from a simple source distribution to a data endpoint, under conditioning. Let $c$ denote the conditioning signal (a binary mask in our case). A state $\mathbf{x}_t$ evolves according to the ODE
\begin{equation}
\frac{d\mathbf{x}_t}{dt} \;=\; \mathbf{u}_\theta(\mathbf{x}_t, t \mid c),
\label{eq:ode}
\end{equation}
so integrating from $t{=}0$ to $1$ maps a seed $\mathbf{x}_0 \!\sim\! p_0$ to an endpoint $\mathbf{x}_1$. Instead of simulating the ODE during training, FM prescribes a reference path that links a noisy source to a noisy data endpoint and exposes a closed-form target velocity. We adopt a straight-line path with constant additive noise,
\begin{equation}
\mathbf{x}_t \;=\; (1{-}t)\,\mathbf{x}_0 \;+\; t\,\mathbf{x}_1 \;+\; \sigma\,\boldsymbol{\epsilon},
\quad \boldsymbol{\epsilon}\!\sim\!\mathcal{N}(\mathbf{0},\mathbf{I}),
\label{eq:interpolant_synflow}
\end{equation}
whose time derivative is $\partial_t \mathbf{x}_t \;=\; \mathbf{x}_1 - \mathbf{x}_0$. The linear term steers the state along the straight line between endpoints, while the additive Gaussian term keeps the path stochastic without affecting the target velocity. Training samples pairs $(\mathbf{x}_0,\mathbf{x}_1,c)$, draws a random $t\in[0,1]$, evaluates $\mathbf{x}_t$ via \eqref{eq:interpolant_synflow}, and regresses $\mathbf{u}_\theta(\mathbf{x}_t,t\mid c)$ to this target velocity. At test time, a deterministic sampler is obtained by numerically integrating \eqref{eq:ode} from the seed. This flow-matching formulation underlies both the SegFlow and SynFlow ODE branches shown in Figure~\ref{fig:overview}.

\subsection{SegFlow: Segmentation as continuous transport}

SegFlow treats image segmentation as a continuous transport problem: instead of classifying pixels independently, it learns a time-indexed velocity field that transports an input image into its ground-truth mask under a deterministic flow (upper branch in Figure~\ref{fig:overview}). Concretely, let $x_0$ be an input image and $x_1$ its binary mask. We define a linear interpolation path from $x_0$ to $x_1$ (cf.\ Eq.~\eqref{eq:interpolant_segflow}), so that for any $t\in[0,1]$ the state $x_t$ lies between the image and mask. The target velocity along this path is
$u_t^{\mathrm{target}} \;=\; x_1 - x_0$, \textit{i.e.}, the constant vector difference from image to mask. We train a U\mbox{-}Net–based neural ODE model $u_\theta(x_t, t)$ to regress this vector field by minimizing the mean-squared error:
\begin{equation}
\label{eq:segflow_loss}
\mathcal{L}_{\mathrm{SegFlow}}
\;=\;
\mathbb{E}_{(x_0,x_1),\,t\sim[0,1]}
\!\left[\,
\big\|\,u_\theta(x_t,t) - (x_1 - x_0)\big\|_2^2
\,\right].
\end{equation}
In practice, we sample $t$ uniformly, compute
\begin{equation}
\label{eq:interpolant_segflow}
x_t \;=\; (1-t)\,x_0 \;+\; t\,x_1,
\end{equation}
then regress $u_\theta(x_t,t)$ to $x_1{-}x_0$; in implementation, we scale time as $\tau = 1000\,t$ for the U-Net time embedding.

Importantly, we do not predict logits or probabilities: at every timestep, the output of the network is a velocity field which receives a target supervision as per Eq.~\ref{eq:segflow_loss}. By supervising the formation of the mask over $t\!\in\![0,1]$, SegFlow implicitly encourages topological consistency (\textit{e.g.}, connected branches) early in the trajectory rather than only at the final mask. In contrast, traditional segmentation losses (cross-entropy or Dice) supervise only the end-state logits and often miss fine connectivity~\cite{cheng2022mask2former}, even when augmented with topology-aware losses such as clDice~\cite{shit2021cldice}. Our velocity-based loss anchors the vector field at each $t$, so the ODE integrator must carry thin structures intact to reach the true mask.

At test time, SegFlow performs deterministic ODE integration from $t{=}0$ to $1$ starting from the input image, using a fixed-step Euler solver with $K$ steps.
Given $x_0$, we iteratively update
\begin{equation}
\label{eq:ode_update}
x_{t+\Delta t}
\;=\;
x_t \;+\; \Delta t\,u_\theta(x_t,t),
\end{equation}
where $\Delta t = 1/K$ and $t \in \{0,\Delta t,2\Delta t,\dots,1-\Delta t\}$, to obtain the
final state $x_1 \!=\! \hat M$. We then threshold $\hat M$ to yield a binary segmentation.

When a thin branch would otherwise disconnect, the velocity field must repair it over $t$ so that $x_1$ matches the ground-truth mask. In practice, SegFlow yields more complete and connected centerlines, preserving topology without auxiliary skeleton losses~\cite{shit2021cldice} or post-hoc graph heuristics.

\noindent\textbf{Self-Conditioning Mechanism.} For ultra-thin structures (e.g., cracks), we additionally use a self-conditioning variant (SegFlow-SC) in which the velocity network also receives a running estimate of the terminal mask. SegFlow-SC replaces the velocity network by
$u_{\theta}^{\mathrm{SC}}([x_t,\tilde{x}_1],t)$, where $\tilde{x}_1$ is an estimate of the
final mask at time $t$ and $[\cdot,\cdot]$ denotes channel-wise concatenation.

Under the straight-line interpolant (Eq.~\eqref{eq:interpolant_segflow}) and constant target velocity $u_t^{\mathrm{target}}$,
a predicted velocity implies an endpoint estimate
\begin{equation}
\hat{x}_1 \;=\; x_t + (1-t)\,u_{\theta}^{\mathrm{SC}}([x_t,\tilde{x}_1],t).
\label{eq:sc_endpoint}
\end{equation}

During training, with probability $p_{\mathrm{sc}}$, we first run the network with
$\tilde{x}_1=\mathbf{0}$ to form $\hat{x}_1$, stop-gradient it, and then run a second pass
conditioned on this estimate:
\begin{equation}
\tilde{x}_1 \leftarrow \mathrm{stopgrad}\!\left(x_t + (1-t)\,u_{\theta}^{\mathrm{SC}}([x_t,\mathbf{0}],t)\right).
\label{eq:sc_train}
\end{equation}
The loss remains the same MSE regression to the FM target velocity, but applied to the
self-conditioned prediction.


\subsection{SynFlow: Rendering as topology-aware transport}
\label{sec:synflow}
SynFlow synthesizes a photorealistic image that is layout- and connectivity-consistent with a given thin-structure mask. The generator is an ODE sampler conditioned on the mask: starting from a noise seed $\mathbf{z}\!\sim\!p_0$, the predicted image is the terminal state of the flow
\begin{equation}
\hat{\mathbf{x}} \;=\; \mathbf{z} \;+\; \int_{0}^{1} \mathbf{u}_\theta(\mathbf{x}_t, t \mid \mathbf{M})\,dt ,
\label{eq:synflow_ode}
\end{equation}
implemented with a fixed-budget explicit ODE solver. In practice, we use the Dormand–Prince (\texttt{dopri5}) method with at most $K{=}200$ function evaluations; for exposition we write a single explicit step as
\begin{equation}
\mathbf{x}_{t+\Delta t} \;=\; \mathbf{x}_t \;+\; \Delta t \,\mathbf{u}_\theta(\mathbf{x}_t, t \mid \mathbf{M}),
\label{eq:euler}
\end{equation}
For each training pair $(\mathbf{x}_1,\mathbf{M})$, SynFlow instantiates the mask-conditioned FM interpolant with $(\mathbf{x}_0,\mathbf{x}_1)\!=\!(\mathbf{z},\mathbf{x}_1)$,
\begin{equation}
\mathbf{x}_t \;=\; (1{-}t)\,\mathbf{z} \;+\; t\,\mathbf{x}_1 \;+\; \sigma\,\boldsymbol{\epsilon},
\quad \boldsymbol{\epsilon}\!\sim\!\mathcal{N}(\mathbf{0},\mathbf{I}),
\label{eq:synflow_interpolant}
\end{equation}
which yields the closed-form target velocity $\partial_t \mathbf{x}_t \;=\; \mathbf{x}_1 - \mathbf{z}$; the training objective regresses the conditional velocity along this path,
\begin{equation}
\begin{aligned}
\mathcal{L}_{\text{SynFlow}}
&=
\mathbb{E}_{(\mathbf{x}_1,\mathbf{M})}\,
\mathbb{E}_{\mathbf{z},\boldsymbol{\epsilon}}\,
\mathbb{E}_{t\sim\mathcal{U}(0,1)}
\\
&\quad
\left[
\big\|\mathbf{u}_\theta(\mathbf{x}_t, t \mid \mathbf{M}) - \partial_t \mathbf{x}_t\big\|_2^2
\right].
\end{aligned}
\label{eq:synflow_loss}
\end{equation}

Training SynFlow is straightforward: we sample $(\mathbf{x}_1,\mathbf{M})$, a noise seed $\mathbf{z}$, and time $t$, then optimize the conditional velocity field with the pathwise objective in Eq.~\eqref{eq:synflow_loss}. This supervision, together with deterministic sampling, encourages long-range connectivity and yields interpretable trajectories in which topology typically stabilizes earlier than texture.

To effectively synthesize connectivity-consistent thin structures, we design a custom UNet-based architecture for SynFlow. Our architecture is driven by three crucial additions in standard UNet that make the conditioning highly effective:

\noindent\textbf{Multi-Scale Mask Injection.} We explicitly supply mask information at multiple matching spatial resolutions throughout the network (while time enters via a sinusoidal or MLP embedding). This ensures that structural geometry is injected and preserved at every scale of the feature hierarchy.

\noindent\textbf{Lite-SPADE GroupNorm (LSGN). }We propose the LSGN module to tightly couple normalization to the mask pyramid (Fig.~\ref{fig:overview}, left). A small convolutional head maps the downsampled mask at each scale to spatial affine parameters that modulate group-normalized features:

\begin{equation}
\mathrm{LSGN}(h \mid \mathbf{M}^{(s)}) \;=\; \gamma(\mathbf{M}^{(s)}) \odot \mathrm{GN}(h) \;+\; \beta(\mathbf{M}^{(s)}).
\label{eq:lsgn}
\end{equation}
This novel mechanism keeps thin structures aligned through pooling and upsampling while remaining parameter-efficient (per-group rather than per-channel modulation). 

\noindent\textbf{Boundary-gating Module (BGM)}. we propose the BGM to prioritize the narrow band where topological errors originate (Fig.~\ref{fig:overview}, right). A soft edge map is derived from the mask using a morphological gradient and passed through a $3{\times}3$ convolution plus sigmoid to obtain a spatial gate; decoder and skip features are then multiplicatively modulated,
\begin{equation}
h_{\text{out}}(\mathbf{p}) \;=\; G_B(\mathbf{p}) \, h_{\text{in}}(\mathbf{p}),
\label{eq:bgm}
\end{equation}
By dynamically allocating capacity to high-curvature interfaces and discourages bleed-through, the BGM ensures the network focuses on the most difficult boundary regions. Together, multi-scale mask conditioning, LSGN, and BGM concentrate modeling power exactly where connectivity is fragile, transforming the baseline UNet into an architecture specialized for thin structures.

\noindent\textbf{Controllable mask generator.}
To steer SynFlow toward regimes where annotation is most expensive, we pair it with a controllable mask generator. For in-domain data, we train a small conditional flow-matching model whose class label encodes crack sparsity bins (e.g., 0–0.5\%, 0.5–1\%); sampling a bin yields a mask that SynFlow renders into an image–mask pair. For cross-domain robustness, we further enrich masks via deterministic morphology: area-controlled propagation, branch-length extension with matched width, and thin/thick variants derived from skeleton radii. These operations expose SynFlow to sparsity, thickness, and branching patterns that are rare or absent in the original datasets.

\section{Experiments}
\label{sec:exp}

\subsection{Experimental settings}
\label{sec:exp_settings}


\noindent\textbf{Datasets.}
We evaluate FMS$^2$ on five standard datasets—three crack sets (CRACK500, CrackTree260, CrackLS315~\cite{zhang2016icip,zou2018deepcrack,zou2012cracktree}) and two vessel sets (DRIVE retinal fundus, XCAD coronary angiography~\cite{staal2004drive,ma2021selfsupervised}).
The crack datasets span handheld to controlled laser acquisition and provide expert binary masks; foreground sparsity is severe, with mean crack coverage of roughly $0.6\%$–$2.5\%$.
DRIVE comprises color fundus photographs with ophthalmologist vessel maps.
XCAD contains coronary X-ray angiograms with curated arterial trees, capturing highly branched vasculature and contrast-flow variability typical of interventional imaging.
Acquisition details, preprocessing, and task conventions are summarized in the supplementary material.



\noindent\textbf{Comparison methods.}
We compare to six recent baselines spanning complementary modeling philosophies for thin structures.
Mask2Former~\cite{cheng2022mask2former} is a general-purpose segmentation transformer, and SemFlow~\cite{wang2024semflow} unifies segmentation and synthesis via latent-space flow matching.
DConnNet~\cite{yang2023dconnnet}, FlowSDF~\cite{flowsdf}, Topograph~\cite{lux2025topograph}, and SCSegamba~\cite{liu2025scsegamba} introduce architecture or topology priors tailored to thin structures, covering transformer-style generalists, rectified/flow objectives, and topology-aware designs for ultra-sparse targets.

\noindent\textbf{Evaluation metrics.}
We report volumetric overlap and topological fidelity.
Volumetric accuracy uses mean IoU and Dice (F1), with
$\mathrm{IoU}=\frac{|\mathcal{P}\cap\mathcal{G}|}{|\mathcal{P}\cup\mathcal{G}|}$ and
$\mathrm{Dice}=\frac{2|\mathcal{P}\cap\mathcal{G}|}{|\mathcal{P}|+|\mathcal{G}|}$,
where $\mathcal{P}$ and $\mathcal{G}$ are predicted and ground-truth foreground sets.
Topology is assessed via clDice~\cite{shit2021cldice} and the Betti matching error $\mu^{\mathrm{err}}_k(\mathcal{P},\mathcal{G})$ for $k\in\{0,1\}$~\cite{stucki2023topologically},
which resolves the limitations of the Betti number error by taking the spatial correspondence of topological features into account.

\noindent\textbf{Implementation details.}
All experiments use PyTorch~2.8.0 on two NVIDIA A100 GPUs with Adam~\cite{kingma2015adam}. Inputs are cropped/resized to $256{\times}256$. For each crack dataset we curate 500 crack-centered patches and split them $80/10/10$ into train/val/test; for DRIVE we use the standard $20/20$ split, and for XCAD we use $100/26$ train/test images. All competing baselines are trained and evaluated using their official code on our data splits. SegFlow is trained for $42$k iterations with batch size $4$ and base learning rate $10^{-5}$, with $5$k-step linear warmup followed by cosine annealing (decay$=0.1$). SynFlow uses the same schedule with batch size $4$, base learning rate $5{\times}10^{-5}$, and $150$k iterations, and is sampled with a deterministic ODE solver using $K{=}10$–$20$ function evaluations at inference. Remaining hyperparameters follow standard practice and are listed in the Appendix.

\subsection{Comparison of SegFlow with SOTA Methods}
\label{sec:comp_sota}

\noindent\textbf{Overall results.}
Table~\ref{tab:main_results} compares SegFlow with six recent baselines across five crack/vessel datasets. SegFlow delivers the best overall performance, improving volumetric overlap and connectivity-aware quality over the strongest prior models. On average, it increases mIoU from 0.511 to 0.599 ($+17.2\%$) and reduces $\mu^{\mathrm{err}}$ from 82.145 to 51.524 ($−37.3\%$). Note that these results are obtained with SegFlow alone, and do not use the SynFlow generated images at all.

\begin{table*}[!htbp]
\centering
\caption{Comparison of SegFlow with 6 SOTA methods across 5 datasets. [Best: \textbf{bold}, Second-best: \underline{underlined}]}
\label{tab:main_results}
\scriptsize
\setlength{\tabcolsep}{3pt}
\renewcommand{\arraystretch}{1.05}
\newcommand{\metrichead}[1]{\raisebox{-0.35ex}{\strut #1}}
\begin{adjustbox}{width=\textwidth,center}
\begin{tabular}{l|l|cc|cc|cc|cc|cc|cc}
\hline
\multicolumn{2}{c|}{} &
\multicolumn{4}{c|}{\textbf{CRACK500}} &
\multicolumn{4}{c|}{\textbf{CrackTree260}} &
\multicolumn{4}{c}{\textbf{CrackLS315}} \\
\hline
\textbf{Method} & \textbf{Venue} &
\multicolumn{2}{c|}{\textbf{Volumetric} (\(\uparrow\))} &
\multicolumn{2}{c|}{\textbf{Topological}} &
\multicolumn{2}{c|}{\textbf{Volumetric} (\(\uparrow\))} &
\multicolumn{2}{c|}{\textbf{Topological}} &
\multicolumn{2}{c|}{\textbf{Volumetric} (\(\uparrow\))} &
\multicolumn{2}{c}{\textbf{Topological}} \\
\cline{3-14}
 &  &
\metrichead{mIoU} & \metrichead{F1} &
\metrichead{clDice\(\uparrow\)} & \metrichead{$\mu^{\mathrm{err}}\downarrow$} &
\metrichead{mIoU} & \metrichead{F1} &
\metrichead{clDice\(\uparrow\)} & \metrichead{$\mu^{\mathrm{err}}\downarrow$} &
\metrichead{mIoU} & \metrichead{F1} &
\metrichead{clDice\(\uparrow\)} & \metrichead{$\mu^{\mathrm{err}}\downarrow$} \\
\hline
Mask2Former   & CVPR~2022    &
0.562 & 0.710 & \underline{0.821} & 27.868 &
0.332 & 0.497 & 0.552 & 119.536 &
0.308 & 0.467 & 0.522 & \textbf{93.310} \\
DConnNet      & CVPR~2023    &
\underline{0.571} & \underline{0.720} & \underline{0.821} & \textbf{23.651} &
0.341 & 0.506 & 0.546 & 252.225 &
0.295 & 0.451 & 0.481 & 120.670 \\
SemFlow       & NeurIPS~2024 &
0.172 & 0.286 & 0.287 & 154.372 &
0.045 & 0.085 & 0.085 & 338.955 &
0.041 & 0.079 & 0.078 & 142.060 \\
FlowSDF       & IJCV~2025    &
0.531 & 0.687 & 0.763 & 36.558 &
\underline{0.577} & \underline{0.728} & \underline{0.729} & \underline{157.830} &
0.221 & 0.358 & 0.358 & 137.150 \\
Topograph     & ICLR~2025    &
0.471 & 0.619 & 0.740 & 43.860 &
0.184 & 0.308 & 0.272 & 161.975 &
0.114 & 0.203 & 0.186 & 180.210 \\
SCSegamba     & CVPR~2025    &
0.549 & 0.702 & 0.794 & \underline{24.791} &
0.437 & 0.604 & 0.609 & 202.800 &
\underline{0.386} & \underline{0.548} & \underline{0.557} & \underline{94.130} \\
\textbf{SegFlow} & \textbf{Ours} &
\textbf{0.612} & \textbf{0.753} & \textbf{0.832} & 24.884 &
\textbf{0.645} & \textbf{0.771} & \textbf{0.772} & \textbf{76.070} &
\textbf{0.442} & \textbf{0.603} & \textbf{0.603} & 96.260 \\
\hline\hline
\multicolumn{2}{c|}{} &
\multicolumn{4}{c|}{\textbf{DRIVE}} &
\multicolumn{4}{c|}{\textbf{XCAD}} &
\multicolumn{4}{c}{\textbf{Mean (5 datasets)}} \\
\hline
\textbf{Method} & \textbf{Venue} &
\multicolumn{2}{c|}{\textbf{Volumetric} (\(\uparrow\))} &
\multicolumn{2}{c|}{\textbf{Topological}} &
\multicolumn{2}{c|}{\textbf{Volumetric} (\(\uparrow\))} &
\multicolumn{2}{c|}{\textbf{Topological}} &
\multicolumn{2}{c|}{\textbf{Volumetric} (\(\uparrow\))} &
\multicolumn{2}{c}{\textbf{Topological}} \\
\cline{3-14}
 &  &
\metrichead{mIoU} & \metrichead{F1} &
\metrichead{clDice\(\uparrow\)} & \metrichead{$\mu^{\mathrm{err}}\downarrow$} &
\metrichead{mIoU} & \metrichead{F1} &
\metrichead{clDice\(\uparrow\)} & \metrichead{$\mu^{\mathrm{err}}\downarrow$} &
\metrichead{mIoU} & \metrichead{F1} &
\metrichead{clDice\(\uparrow\)} & \metrichead{$\mu^{\mathrm{err}}\downarrow$} \\
\hline
Mask2Former   & CVPR~2022 &
0.444 & 0.615 & 0.661 & 280.480 &
0.616 & 0.761 & 0.812 & 15.923 &
0.452 & 0.610 & 0.674 & 107.423 \\
DConnNet      & CVPR~2023 &
0.567 & 0.722 & \underline{0.780} & \textbf{48.104} &
\underline{0.636} & \underline{0.776} & \textbf{0.835} & \textbf{9.827} &
0.482 & 0.635 & 0.693 & 90.895 \\
SemFlow       & NeurIPS~2024 &
0.295 & 0.454 & 0.449 & 149.208 &
0.157 & 0.269 & 0.225 & 148.942 &
0.142 & 0.235 & 0.225 & 186.707 \\
FlowSDF       & IJCV~2025 &
\underline{0.635} & \underline{0.776} & 0.776 & 130.313 &
0.592 & 0.742 & 0.780 & 18.125 &
\underline{0.511} & 0.658 & 0.681 & 95.995 \\
Topograph     & ICLR~2025 &
0.553 & 0.711 & 0.676 & 239.062 &
0.536 & 0.696 & 0.676 & 120.500 &
0.372 & 0.507 & 0.510 & 149.121 \\
SCSegamba     & CVPR~2025 &
0.605 & 0.753 & 0.765 & 65.563 &
0.564 & 0.720 & 0.759 & 23.442 &
0.508 & \underline{0.665} & \underline{0.697} & \underline{82.145} \\
\textbf{SegFlow} & \textbf{Ours} &
\textbf{0.653} & \textbf{0.789} & \textbf{0.830} & \underline{49.792} &
\textbf{0.642} & \textbf{0.780} & \underline{0.831} & \underline{10.615} &
\textbf{0.599} & \textbf{0.739} & \textbf{0.774} & \textbf{51.524} \\
\hline
\end{tabular}
\end{adjustbox}
\end{table*}

On the crack datasets, SegFlow yields clear gains in both overlap and topology; for instance, on CrackTree260 it reduces $\mu^{\mathrm{err}}$ from 157.830 to 76.070, indicating substantially fewer topology mismatches in complex branching cracks. On vessel datasets, SegFlow remains highly competitive and improves connectivity-aware quality; for example, on DRIVE it increases clDice from 0.780 to 0.830, reflecting more continuous vessel trees with fewer breaks.

\noindent\textbf{Latent-space unified generative models vs.\ image-space transport.}
Unified segmentation--generation frameworks such as SemFlow~\cite{wang2024semflow} learn a latent-space flow that must simultaneously preserve geometry and model appearance. On ultra-sparse, one-pixel targets, this coupling is brittle and often washes out thin structures, leading to low overlap and large topological mismatch (high $\mu^{\mathrm{err}}$ in Table~\ref{tab:main_results}). In contrast, SegFlow performs explicit \emph{image-space} transport from image to mask with trajectory-level supervision, so connectivity is enforced throughout the ODE evolution rather than only at the final prediction. This difference is especially pronounced on topology-fragile crack data such as CrackTree260.


\noindent\textbf{Effect of training loss on topological fidelity.}
While architectural advances are a major direction for improving thin-structure segmentation (Table~\ref{tab:main_results}), topology-aware \emph{loss design} is another effective line of work. Table~\ref{tab:segflow_loss_variants_muerr} isolates the role of supervision by training an identical U\mbox{-}Net backbone (matching SegFlow’s architecture) with different objectives, including clCE and Skeleton Recall, and evaluating topology using $\mu^{\mathrm{err}}$. While topology-aware end-state losses reduce some failure modes, they still supervise only the final prediction. In contrast, SegFlow’s flow-matching objective supervises the entire transport trajectory, which consistently yields lower $\mu^{\mathrm{err}}$ across datasets and improves the mean score. Although clCE improves the mean $\mu^{\mathrm{err}}$ from 82.145 (SCSegamba in Table~\ref{tab:main_results}) to 71.348, it still leaves a large gap to SegFlow’s 51.524, suggesting that trajectory-level supervision is a stronger inductive bias for preserving thin-structure connectivity.

\begin{table*}[t]
\centering
\caption{$\mu^{\mathrm{err}}$ comparison among SegFlow variants across five datasets (lower is better). Best results are in bold, and second-best results are underlined.}
\label{tab:segflow_loss_variants_muerr}
\setlength{\tabcolsep}{6pt}
\renewcommand{\arraystretch}{1.05}
\begin{adjustbox}{width=\textwidth}
\begin{tabular}{@{} l l c c c c c c @{}}
\toprule
\textbf{Method} & \textbf{Venue} &
\textbf{CRACK500} & \textbf{CrackTree260} & \textbf{CrackLS315} & \textbf{DRIVE} & \textbf{XCAD} & \textbf{Mean} \\
\midrule
clCE & MICCAI &
\underline{25.163} & 166.415 & \textbf{92.700} & \underline{55.479} & \underline{16.982} & \underline{71.348} \\
Skeleton Recall & ECCV~2024 &
26.779 & \underline{144.850} & \underline{92.930} & 85.898 & 18.231 & 73.738 \\
\textbf{SegFlow} & \textbf{Ours} &
\textbf{24.884} & \textbf{76.070} & 96.260 & \textbf{49.792} & \textbf{10.615} & \textbf{51.524} \\
\bottomrule
\end{tabular}
\end{adjustbox}
\end{table*}

\begin{figure}[!htbp]
  \centering
  \includegraphics[width=1\linewidth]{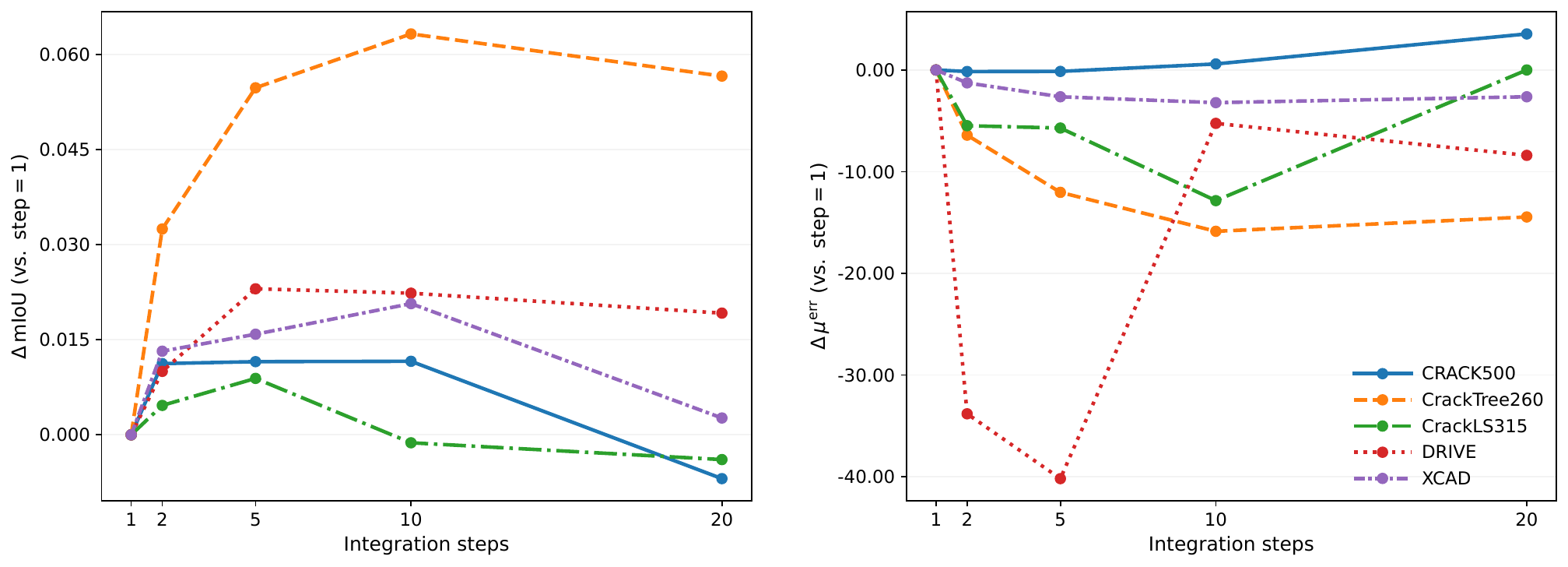}
  \caption{Impact of integration steps in SegFlow on segmentation performance. We report the change (relative to step=1) in mIoU and $\mu^{\mathrm{err}}$ across datasets as the number of integration steps varies.}
  \label{fig:segflow_integration_steps}
\end{figure}

\noindent\textbf{Number of integration steps.}
Figure~\ref{fig:segflow_integration_steps} studies the effect of SegFlow's Euler integration budget. Too few steps cause underfitting and under-developed masks, leading to reduced overlap and higher topological error. Increasing the steps improves both mIoU and $\mu^{\mathrm{err}}$ until performance saturates. This demonstrates that SegFlow is robust to precise step tuning beyond the low-budget regime, motivating our fixed, small-step budget to balance accuracy and compute.


\noindent\textbf{Self-conditioning.}
Table~\ref{tab:segflow_self_conditioning_ablation_rows} evaluates an optional self-conditioning input that feeds a running terminal-mask estimate back into the velocity network. Across all five datasets, this improves mIoU and reduces $\mu^{\mathrm{err}}$, indicating fewer topology-breaking artifacts during transport. These gains are most pronounced on sparse crack datasets where minor discontinuities quickly fragment centerlines. As a lightweight modification, self-conditioning provides a practical way to stabilize ultra-thin structures with minimal complexity.


\begin{table*}[t]
  \centering
  \caption{Ablation of self-conditioning in SegFlow across five datasets.}
  \label{tab:segflow_self_conditioning_ablation_rows}
  \setlength{\tabcolsep}{10pt}
  \renewcommand{\arraystretch}{1.1}
  \begin{adjustbox}{width=\textwidth}
  \begin{tabular}{@{}lccccc@{}}
    \toprule
    Metric / Variant & CRACK500 & CrackTree260 & CrackLS315 & DRIVE & XCAD \\
    \midrule
    $\uparrow~$mIoU (w/o Self-Cond.) & 0.612 & 0.645 & 0.418 & 0.653 & 0.642 \\
    \phantom{$\uparrow~$mIoU} (w/ Self-Cond.) & \textbf{0.628} & \textbf{0.654} & \textbf{0.453} & \textbf{0.664} & \textbf{0.650} \\
    \midrule
    $\downarrow~\mu^{\mathrm{err}}$ (w/o Self-Cond.) & 24.88 & 76.07 & 96.26 & 49.79 & 10.62 \\
    \phantom{$\downarrow~\mu^{\mathrm{err}}$} (w/ Self-Cond.) & \textbf{21.26} & \textbf{73.04} & \textbf{92.91} & \textbf{41.68} & \textbf{9.68} \\
    \bottomrule
  \end{tabular}
  \end{adjustbox}
\end{table*}

\begin{figure*}[!htbp]
  \centering
  \includegraphics[width=1\textwidth]{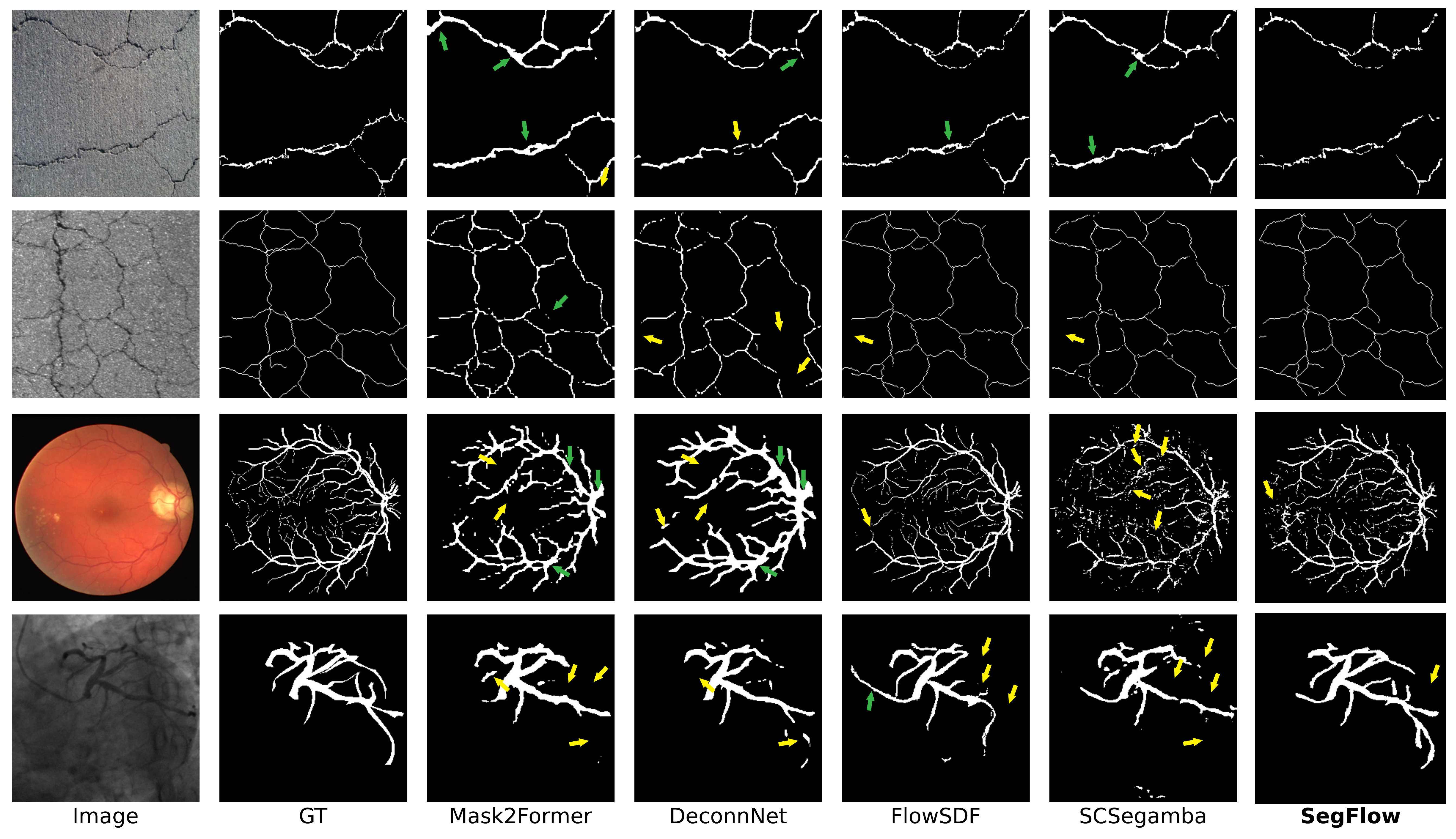}
  \caption{\textbf{Visual comparison of SegFlow against SOTA methods.} Yellow arrows denote missed structures (FNs), green arrows indicate spurious predictions (FPs). \vspace{-1ex}}

  \label{fig:vis_comp}
\end{figure*}

\noindent\textbf{Qualitative comparison.}
We qualitatively analyze SegFlow in Figure~\ref{fig:vis_comp} on representative crack and vessel examples.
Across the crack rows, competing methods either miss thin side branches or introduce spurious fragments (yellow and green arrows), whereas SegFlow recovers nearly all annotated branches with clean, one-pixel traces and intact junctions.
On DRIVE and XCAD, transformer- and Mamba-based methods tend to break peripheral vessels or hallucinate wisps around dense trees, while SegFlow yields coherent arterial networks that follow vessel curvature more faithfully.
We omit SemFlow and Topograph from Figure~\ref{fig:vis_comp}, as their predictions collapse on the thinnest crack datasets (Topograph on CRACK500/CrackLS315, SemFlow on all five datasets; see Table~\ref{tab:main_results}); additional qualitative examples are provided in the Appendix.

\subsection{Combining SegFlow with SynFlow}
\noindent\textbf{Label efficiency with partial real supervision. }Before using SynFlow for augmentation, we evaluate the fidelity of generated masks and SynFlow-rendered, mask-conditioned images using FID (Table~\ref{tab:mask_image_fid}), computed for real vs.\ synthetic masks and real vs.\ SynFlow-rendered images. Figure~\ref{fig:fid} provides qualitative side-by-side examples (more in the supplementary material). Low FID suggests that SynFlow produces realistic, pixel-aligned pairs, supporting its use to reduce annotation cost and improve robustness under domain shift.

\begin{figure*}[t]
  \centering
  \includegraphics[width=0.85\linewidth]{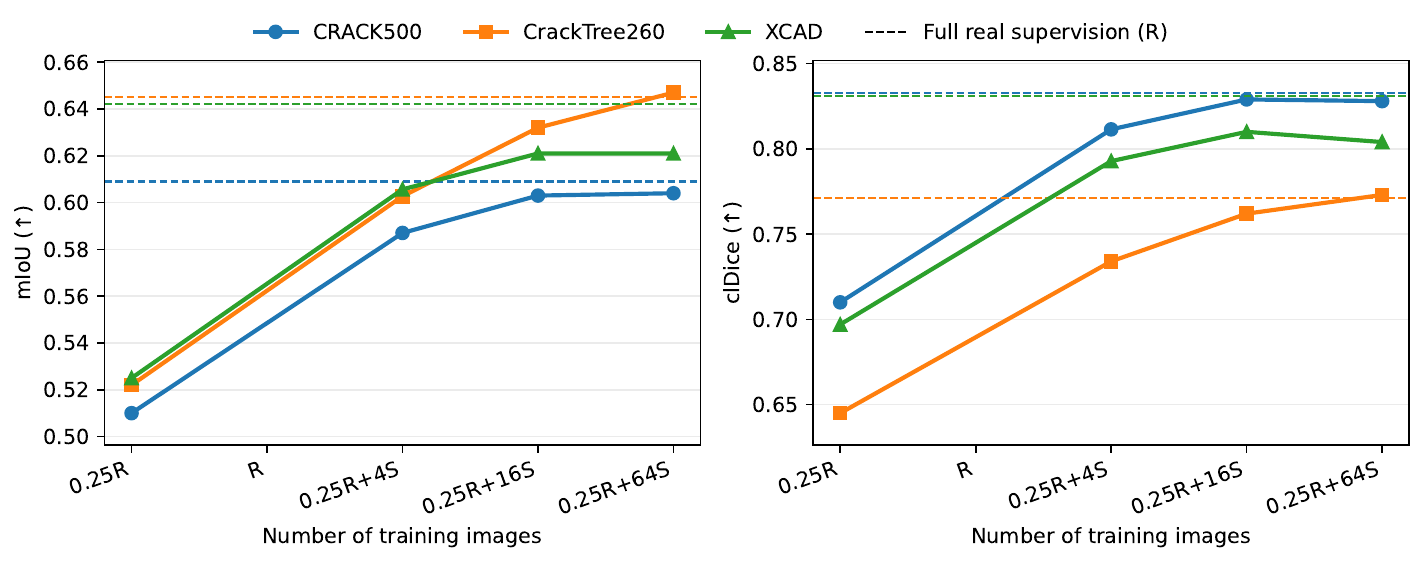}
  \caption{
  \textbf{Effect of SynFlow synthesis on label efficiency.} 
  SegFlow performance when training with a small portion of real labels (\(0.25R\)) and progressively adding SynFlow images (\(0.25R{+}4S\), \(0.25R{+}16S\), \(0.25R{+}64S\)). 
  Here \(R\) is the full real training set (\(R{=}400\) for Crack500 and CrackTree260, \(R{=}100\) for XCAD) and \(S{=0.25}R\); the dashed line marks full real supervision at \(R\). 
  Left: IoU, right: clDice. }
  \label{fig:synflow_label_eff}
\end{figure*}

Figure~\ref{fig:synflow_label_eff} studies how SynFlow trades synthetic images for expert annotations. We train SegFlow with only \(0.25R\) labeled images (where \(R\) is the full real training set) and progressively add SynFlow-generated, pixel-aligned image--mask pairs whose masks are sampled from the controllable mask generator to target diverse sparsity and topology regimes. With \(0.25R\) alone, both IoU and clDice drop noticeably compared to full supervision \(R\). As more synthetic pairs are added (\(0.25R{+}4S\rightarrow0.25R{+}16S\rightarrow0.25R{+}64S\)), performance improves consistently and, on the crack datasets, largely closes the gap to \(R\). On XCAD, SynFlow also yields clear gains over \(0.25R\), although a residual gap to full supervision remains, reflecting the higher appearance variability of coronary angiography. Overall, these trends indicate that SynFlow can substantially reduce the amount of expensive pixel-level annotation needed to reach strong performance, making it a practical data-centric tool for label-efficient thin-structure segmentation.

  \begin{table}[t]
  \centering
  \caption{FID of generated masks and the corresponding mask-conditioned images generated by SynFlow (Mask / Image; lower is better).}
  \label{tab:mask_image_fid}
  \setlength{\tabcolsep}{3pt}
  \renewcommand{\arraystretch}{1.0}
  \begin{adjustbox}{max width=\columnwidth}
  \begin{tabular}{@{}lccccc@{}}
    \toprule
    Metric & CRACK500 & CrackTree260 & CrackLS315 & DRIVE & XCAD \\
    \midrule
    FID (Mask / Image) &
    25.67 / 18.36 &
    15.68 / 20.14 &
    13.84 / 19.81 &
    15.04 / 3.04 &
    76.02 / 40.13 \\
    \bottomrule
  \end{tabular}
  \end{adjustbox}
\end{table}

\begin{figure*}[t]
  \centering
  \includegraphics[width=\textwidth]{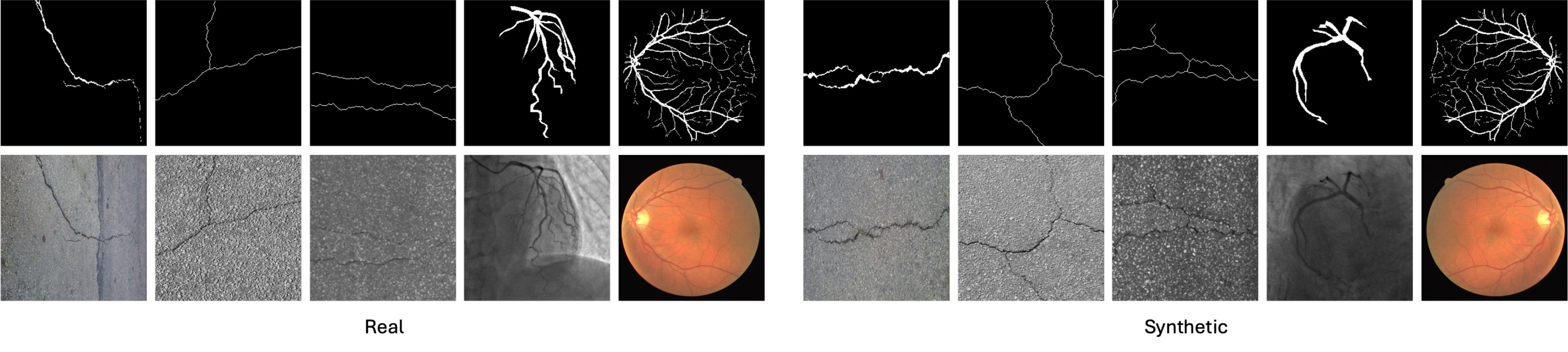}
  \vspace{-1.5em}
  \caption{Comparison of real vs.\ SynFlow-generated mask--image pairs. SynFlow preserves pixel alignment between masks and rendered images while producing realistic textures; additional examples are provided in the supplementary material.}
  \label{fig:fid}
\end{figure*}

\begin{table*}[!htbp]
\centering
\caption{Cross-domain generalization across three crack datasets. Each cell reports mIoU/F1. Values in parentheses indicate relative gains (\%) over SegFlow.}
\label{tab:cross_domain_vol_only}
\setlength{\tabcolsep}{6pt}
\renewcommand{\arraystretch}{1.08}
\begin{adjustbox}{width=\textwidth,center}
\begin{tabular}{l c c c}
\toprule
\textbf{Method} &
\makecell{\textbf{CRACK500$\rightarrow$}\\\textbf{CrackTree260}} &
\makecell{\textbf{CrackLS315B$\rightarrow$}\\\textbf{CRACK500}} &
\makecell{\textbf{CrackTree260$\rightarrow$}\\\textbf{CRACK500}} \\
\midrule
SCSegamba &
$0.224/0.363$ & $0.212/0.339$ & $0.246/0.384$ \\
FlowSDF &
$0.212/0.347$ & $0.195/0.320$ & $0.247/0.386$ \\
SegFlow &
$0.251/0.396$ & $0.223/0.352$ & $0.254/0.393$ \\
\midrule
SegFlow+DA &
\makecell{$0.259/0.408$ (+3.2\%/+3.0\%)} &
\makecell{$0.229/0.358$ (+2.7\%/+1.7\%)} &
\makecell{$0.257/0.395$ (+1.2\%/+0.5\%)} \\
\textbf{SegFlow+SynFlow} &
\makecell{$0.305/0.444$ (+21.5\%/+12.1\%)} &
\makecell{$0.411/0.566$ (+84.3\%/+60.8\%)} &
\makecell{$0.340/0.489$ (+33.9\%/+24.4\%)} \\
\bottomrule
\end{tabular}
\end{adjustbox}
\end{table*}

\noindent\textbf{Cross-domain robustness under domain shift.}
Table~\ref{tab:cross_domain_vol_only} evaluates source-only transfer across the three crack datasets, where changes in background texture, noise, and acquisition conditions often fragment or erase thin structures and lead to substantial generalization drops. Although SegFlow already outperforms the strongest architecture baselines in-domain (Table~\ref{tab:main_results}), performance still degrades under domain shift. Classic DA (i.e., horizontal/vertical flips, rotations, cropping, brightness--contrast jitter, Gaussian noise, and Gaussian blur) provides only modest benefits, improving mIoU by about $+0.006$ ($+2.4\%$ relative gain) on average across the three cross-domain experiments. In contrast, SynFlow yields substantially larger improvements, increasing mIoU by $+0.109$ on average across the same three experiments, corresponding to about $+46.6\%$ relative gain. This gap is expected because SynFlow is distinct from standard DA. Rather than enforcing invariance via label-preserving transforms, SynFlow is a paired mask$\rightarrow$image generator that produces \emph{pixel-aligned} image--mask samples. Using controllable mask synthesis that varies sparsity, width, and connectivity, together with the renderer's ability to model realistic texture and appearance variation conditioned on these masks, it introduces structured, topology-relevant shifts in the source domain that better match cross-domain failure modes.

\vspace{-0.9em}
\section{Conclusion}
\label{sec:con}
\vspace{-0.5em}
We present FMS$^2$, a dual-module flow-matching framework for thin-structure learning. Its first module, SegFlow, performs deterministic image-to-mask transport with trajectory-level supervision, consistently outperforming strong baselines across five datasets (improving mIoU from 0.511 to 0.599 and reducing $\mu^{\mathrm{err}}$  from 82.145 to 51.524). The complementary SynFlow module is a controllable mask-to-image renderer that generates realistic training pairs. Using SynFlow for data augmentation reduces manual annotation needs by 75\% while recovering fully supervised performance and substantially improving cross-domain robustness. Finally, to support future benchmarking, we will release a large-scale dataset of 10k crack and 1k vessel image-mask pairs.

%
%
\bibliographystyle{splncs04}
\bibliography{main}

@String(CVPR  = {IEEE Conf. Comput. Vis. Pattern Recog.})

@String(ICCV  = {Int. Conf. Comput. Vis.})

@String(ECCV  = {Eur. Conf. Comput. Vis.})

@String(NeurIPS = {Adv. Neural Inform. Process. Syst.})

@String(ICLR  = {Int. Conf. Learn. Represent.})

@String(AAAI  = {AAAI})

@String(ICIP  = {IEEE Int. Conf. Image Process.})

@String(CVPR  = {CVPR})

@String(ICCV  = {ICCV})

@String(ECCV  = {ECCV})

@String(NeurIPS = {NeurIPS})

@String(ICLR  = {ICLR})

@String(ICIP  = {ICIP})

@article{mou2021cs2net,
  author = {Mou, Lei and Zhao, Yitian and Fu, Huazhu and Liu, Yonghuai and Cheng, Jun and Zheng, Yalin and Su, Pan and Yang, Jianlong and Chen, Li and Frangi, Alejandro F. and Akiba, Masahiro and Liu, Jiang},
  title = {{CS2-Net}: Deep learning segmentation of curvilinear structures in medical imaging},
  journal = {Medical Image Analysis},
  volume = {67},
  pages = {101874},
  year = {2021},
  doi = {10.1016/j.media.2020.101874}
}

@inproceedings{shit2021cldice,
  author = {Shit, Suprosanna and Paetzold, Johannes C. and Sekuboyina, Anjany and Ezhov, Ivan and Unger, Alexander and Zhylka, Andrey and Pluim, Josien P. W. and Bauer, Ulrich and Menze, Bjoern H.},
  title = {clDice: A Novel Topology-Preserving Loss Function for Tubular Structure Segmentation},
  booktitle = {Proceedings of the IEEE/CVF Conference on Computer Vision and Pattern Recognition (CVPR)},
  year = {2021},
  pages = {16560--16569},
  doi = {10.1109/CVPR46437.2021.01629}
}

@inproceedings{yang2023dconnnet,
  author = {Yang, Ziyun and Farsiu, Sina},
  title = {{Directional Connectivity-based Segmentation of Medical Images}},
  booktitle = {Proceedings of the IEEE/CVF Conference on Computer Vision and Pattern Recognition (CVPR)},
  year = {2023},
  pages = {11525--11535},
  doi = {10.1109/CVPR.2023.01167}
}

@article{gupta2023topologyuncertainty,
  author = {Gupta, Saumya and Zhang, Yikai and Hu, Xiaoling and Prasanna, Prateek and Chen, Chao},
  title = {{Topology-Aware Uncertainty for Image Segmentation}},
  journal = {Advances in Neural Information Processing Systems (NeurIPS)},
  volume = {36},
  year = {2023}
}

@article{zuo2024crackmamba,
  author = {Zuo, Xin and Sheng, Yu and Shen, Jifeng and Shan, Yongwei},
  title = {{Topology-aware Mamba for Crack Segmentation in Structures}},
  journal = {Automation in Construction},
  volume = {168, Part A},
  pages = {105845},
  year = {2024},
  doi = {10.1016/j.autcon.2023.105845}
}

@article{lesage2009review,
  author = {Lesage, David and Angelini, Elsa D. and Bloch, Isabelle and Funka-Lea, Gareth},
  title = {{A review of 3D vessel lumen segmentation techniques: models, features and extraction schemes}},
  journal = {Medical Image Analysis},
  volume = {13},
  number = {6},
  pages = {819--845},
  year = {2009},
  doi = {10.1016/j.media.2009.07.011}
}

@article{zhao2021retvesselreview,
  author = {Chen, Chunhui and Chuah, Joon Huang and Ali, Radzi and Wang, Yizhou},
  title = {{Retinal Vessel Segmentation Using Deep Learning: A Review}},
  journal = {IEEE Access},
  volume = {9},
  pages = {111985--112004},
  year = {2021},
  doi = {10.1109/ACCESS.2021.3102176}
}

@article{rajitha2023retinalreview,
  author = {Rajitha, K. V. and Prasad, Keerthana and Yegneswaran, Prakash P.},
  title = {{Segmentation and Classification Approaches of Clinically Relevant Curvilinear Structures: A Review}},
  journal = {Journal of Medical Systems},
  volume = {47},
  number = {1},
  pages = {40},
  year = {2023},
  doi = {10.1007/s10916-023-01927-2}
}

@article{zhou2023crackreview,
  author = {Zhou, Shanglian and Canchila, Cristia and Song, Wei},
  title = {{Deep learning-based crack segmentation for civil infrastructure: data types, architectures, and benchmarked performance}},
  journal = {Automation in Construction},
  volume = {146},
  pages = {104623},
  year = {2023},
  doi = {10.1016/j.autcon.2023.104623}
}

@article{wu2025crackreview,
  author = {Shi, Yaolong and **(others)**},
  title = {{Deep Learning in Crack Detection: A Comprehensive Review}}, 
  journal = {*** (in press, 2025)***},
  year = {2025}
}

@article{isensee2018nnunet,
  title   = {nnU-Net: a self-configuring method for deep learning-based biomedical image segmentation},
  author  = {Isensee, Fabian and J{\"a}ger, Paul F and Kohl, Simon AA and Petersen, Jens and Maier-Hein, Klaus H},
  journal = {Nature Methods},
  volume  = {18},
  number  = {2},
  pages   = {203--211},
  year    = {2021}
}

@article{chen2024transunet,
  author = {Chen, Jieneng and Mei, Jieru and **(others)** and Zhou, Yuyin},
  title = {{TransUNet: Rethinking the U-Net architecture design for medical image segmentation through the lens of transformers}},
  journal = {Medical Image Analysis},
  volume = {97},
  pages = {103280},
  year = {2024},
  doi = {10.1016/j.media.2023.103280}
}

@inproceedings{wu2024medsegdiffv2,
  author = {Wu, Haohan and Li, Wenxuan and Ye, Yiwen and Tang, Yucheng and Chen, Jing and **(others)**},
  title = {{MedSegDiff-V2: Diffusion-Based Medical Image Segmentation with Transformer}},
  booktitle = {Proceedings of the AAAI Conference on Artificial Intelligence (AAAI)},
  year = {2024}
}

@inproceedings{wu2025denver,
  author = {Wu, Chun-Hung and Chen, Shih-Hong and Hu, Chih-Yao and Wu, Hsin-Yu and Chen, Kai-Hsin and Chen, Yu-You and Su, Chih-Hai and Lee, Chih-Kuo and Liu, Yu-Lun},
  title = {{DeNVeR: Deformable Neural Vessel Representations for Unsupervised Video Vessel Segmentation}},
  booktitle = {Proceedings of the IEEE/CVF Conference on Computer Vision and Pattern Recognition (CVPR)},
  year = {2025}
}

@inproceedings{acebes2024clce,
  author = {Acebes, Cesar and Moustafa, Abdel Hakim and Camara, Oscar and Galdran, Adrian},
  title = {{The Centerline-Cross Entropy Loss for Vessel-Like Structure Segmentation: Better Topology Consistency Without Sacrificing Accuracy}},
  booktitle = {Proceedings of Medical Image Computing and Computer Assisted Intervention (MICCAI) 2024},
  year = {2024},
  pages = {710--720},
  doi = {10.1007/978-3-031-72111-3\_67}
}

@article{xu2023dcsau,
  author = {Xu, Qing and Ma, Zhicheng and He, Na and Duan, Wenting},
  title = {{DCSAU-Net: A deeper and more compact split-attention U-Net for medical image segmentation}},
  journal = {Computers in Biology and Medicine},
  volume = {154},
  pages = {106626},
  year = {2023},
  doi = {10.1016/j.compbiomed.2023.106626}
}

@article{ouyang2025hybridvessel,
  author = {Ouyang, Yihao and Kuang, Xunheng and Xiong, Mengjia and Wang, Zhida and Wang, Yuanquan},
  title = {{A Novel Hybrid Approach for Retinal Vessel Segmentation with Dynamic Long-Range Dependency and Multi-Scale Retinal Edge Fusion Enhancement}},
  journal = {Pattern Recognition},
  year = {2025},
  doi = {10.1016/j.patcog.2024.110387}
}

@inproceedings{zhou2025nnwnet,
  author = {Zhou, Yanfeng and Li, Lingrui and Che, Zhengbang},
  title = {{nnWNet: Rethinking the Use of Transformers in Biomedical Image Segmentation and Calling for a Unified Evaluation Benchmark}},
  booktitle = {Proceedings of the IEEE/CVF Conference on Computer Vision and Pattern Recognition (CVPR)},
  year = {2025},
  pages = {20852--20862}
}

@inproceedings{liu2025scsegamba,
  author    = {Liu, Hui and Jia, Chen and Shi, Fan and Xu, Cheng and Chen, Shengyong},
  title     = {SCSegamba: Lightweight Structure-Aware Vision Mamba for Crack Segmentation in Structures},
  booktitle = {Proceedings of the IEEE/CVF Conference on Computer Vision and Pattern Recognition (CVPR)},
  year      = {2025}
}

@article{liu2024mambahrnet,
  author = {Liu, Jie and Li, Deyuan and Xu, Xin and **(others)**},
  title = {{MambaHRNet: Enhanced High-Resolution Network for Bridge Damage Segmentation}},
  journal = {*submitted*},
  year = {2024}
}

@article{song2025mambafuse,
  author = {Song, Yunlong and Kang, Jian and Su, Yang and Zhang, Shikai and Zhang, Qi and Yu, Yifeng and Zhan, Zhaoxiang and Zhang, Wei},
  title = {{MambaFuse: Cross-scale state space fusion for crack segmentation}},
  journal = {Developments in the Built Environment},
  volume = {24},
  pages = {100751},
  year = {2025},
  doi = {10.1016/j.dibe.2023.100751}
}

@inproceedings{cheng2022mask2former,
  author = {Cheng, Bowen and Misra, Ishan and Schwing, Alexander G. and Kirillov, Alexander},
  title = {{Masked-Attention Mask Transformer for Universal Image Segmentation}},
  booktitle = {Proceedings of the IEEE/CVF Conference on Computer Vision and Pattern Recognition (CVPR)},
  year = {2022},
  pages = {1290--1299},
  doi = {10.1109/CVPR52688.2022.00136}
}

@inproceedings{kirchhoff2024skeletonrecall,
  author = {Kirchhoff, Yannick and Heinrich, Mattias P.},
  title = {{Skeleton Recall Loss for Connectivity Conserving and Resource Efficient Segmentation of Thin Tubular Structures}},
  booktitle = {Proceedings of the European Conference on Computer Vision (ECCV)},
  year = {2024},
  pages = {218--234},
  doi = {10.1007/978-3-031-41606-4\_13}
}

@inproceedings{dong2025codice,
  author = {Dong, Shuyue and Zhu, Xuan and Chen, Haiyang and Zheng, Yalin and Frangi, Alejandro F.},
  title = {{coDice: Connectivity-Preserving Dice Loss for 2D/3D Tubular Structure Segmentation}},
  booktitle = {SPIE Medical Imaging},
  volume = {12529},
  pages = {1252918},
  year = {2025},
  doi = {10.1117/12.2651373}
}

@inproceedings{lux2025topograph,
  author    = {Lux, Laurin and Berger, Alexander H. and Weers, Alexander and Stucki, Nico and R{\"u}ckert, Daniel and Bauer, Ulrich and Paetzold, Johannes C.},
  title     = {Topograph: An Efficient Graph-Based Framework for Strictly Topology Preserving Image Segmentation},
  booktitle = {International Conference on Learning Representations (ICLR)},
  year      = {2025}
}

@inproceedings{rombach2022ldm,
  author = {Rombach, Robin and Blattmann, Andreas and Lorenz, Dominik and Esser, Patrick and Ommer, Bj{\"o}rn},
  title = {{High-Resolution Image Synthesis with Latent Diffusion Models}},
  booktitle = {Proceedings of the IEEE/CVF Conference on Computer Vision and Pattern Recognition (CVPR)},
  year = {2022},
  pages = {10684--10695},
  doi = {10.1109/CVPR52688.2022.01046}
}

@inproceedings{wang2024semflow,
  author = {Wang, Chaoyang and Li, Xiangtai and Qi, Lu and Ding, Henghui and Tong, Yunhai and Yang, Ming-Hsuan},
  title = {{SemFlow: Binding Semantic Segmentation and Image Synthesis via Rectified Flow}},
  booktitle = {Advances in Neural Information Processing Systems (NeurIPS)},
  year = {2024}
}

@inproceedings{qi2024unigs,
  author = {Qi, Lu and Yang, Lehan and Guo, Weidong and Xu, Yu and Du, Bo and Jampani, Varun and Yang, Ming-Hsuan},
  title = {{UniGS: Unified Representation for Image Generation and Segmentation}},
  booktitle = {Proceedings of the IEEE/CVF Conference on Computer Vision and Pattern Recognition (CVPR)},
  year = {2024},
  pages = {6305--6315},
  doi = {10.1109/CVPR.2024.00638}
}

@inproceedings{zhang2023controlnet,
  author = {Zhang, Lvmin and Rao, Anyi and Agrawala, Maneesh},
  title = {{Adding Conditional Control to Text-to-Image Diffusion Models}},
  booktitle = {Proceedings of the IEEE/CVF International Conference on Computer Vision (ICCV)},
  year = {2023},
  pages = {3836--3847},
  doi = {10.1109/ICCV.2023.03885}
}

@inproceedings{feng2025controlnetxs,
  title     = {ControlNet-XS: Spotting Shortcut Learning in Controlled Image Generation},
  author    = {Zavadski, Roman and Feiden, Wolfgang and Rother, Carsten},
  booktitle = {European Conference on Computer Vision (ECCV)},
  year      = {2024},
  pages     = {4511--4522}
}

@article{yang2025crackdiffusion,
  author = {Song, Yunlong and Su, Yumeng and Zhang, Shiying and Wang, Ruilin and Yu, Youling and Zhang, Weiping and Zhang, Qi},
  title = {{CrackdiffNet: A Novel Diffusion Model for Crack Segmentation and Scale-Based Analysis}},
  journal = {Buildings},
  volume = {15},
  number = {11},
  pages = {1872},
  year = {2025},
  doi = {10.3390/buildings15111872}
}

@inproceedings{ronneberger2015unet,
  title     = {U-Net: Convolutional Networks for Biomedical Image Segmentation},
  author    = {Ronneberger, Olaf and Fischer, Philipp and Brox, Thomas},
  booktitle = {Medical Image Computing and Computer-Assisted Intervention (MICCAI)},
  series    = {Lecture Notes in Computer Science},
  volume    = {9351},
  pages     = {234--241},
  year      = {2015},
  publisher = {Springer}
}

@inproceedings{vangansbeke2024ldmseg,
  title     = {A Simple Latent Diffusion Approach for Panoptic Segmentation and Mask Inpainting},
  author    = {Van Gansbeke, S{\'e}bastien and De Brabandere, Bert},
  booktitle = {Computer Vision -- ECCV 2024},
  series    = {Lecture Notes in Computer Science},
  pages     = {271--290},
  year      = {2024},
  publisher = {Springer}
}

@article{lipman2022flowmatching,
  title   = {Flow Matching for Generative Modeling},
  author  = {Lipman, Yaron and Chen, Ricky T. Q. and Ben-Hamu, Heli and Nickel, Maximilian and Le, Matthew},
  journal = {arXiv preprint arXiv:2210.02747},
  year    = {2022}
}

@inproceedings{liu2023rectifiedflow,
  title     = {Flow Straight and Fast: Learning to Generate and Transfer Data with Rectified Flow},
  author    = {Liu, Yang and Gong, Chence and Kong, Lingqi and Wang, Qiang and Lu, Jian},
  booktitle = {International Conference on Learning Representations (ICLR)},
  year      = {2023}
}

@inproceedings{schusterbauer2025diff2flow,
  title     = {Diff2Flow: Training Flow Matching Models via Diffusion Model Alignment},
  author    = {Schusterbauer, Johannes and Gui, Ming and Fundel, Frank and Ommer, Bj{\"o}rn},
  booktitle = {Proceedings of the IEEE/CVF Conference on Computer Vision and Pattern Recognition (CVPR)},
  year      = {2025}
}

@misc{chen2024flowmatchingtutorial,
  title        = {Flow Matching for Generative Modeling (NeurIPS 2024 Tutorial)},
  author       = {Chen, Ricky T. Q. and Lipman, Yaron and Ben-Hamu, Heli},
  howpublished = {\url{https://aisecure.github.io/flow-matching-tutorial/}},
  year         = {2024},
  note         = {Tutorial website; accessed 2025-11-13}
}

@inproceedings{zhang2024ictm,
  title     = {Flow Priors for Linear Inverse Problems via Iterative Corrupted Trajectory Matching},
  author    = {Zhang, Yasi and Yu, Peiyu and Zhu, Yaxuan and Chang, Yingshan and Gao, Feng and Wu, Ying Nian and Leong, Oscar},
  booktitle = {Advances in Neural Information Processing Systems (NeurIPS)},
  year      = {2024}
}

@inproceedings{yang2024scpcontrolnet,
  title     = {Enriching Information and Preserving Semantic Consistency in Expanding Curvilinear Object Segmentation Datasets},
  author    = {Lei, Qin and Zhong, Jiang and Dai, Qizhu and Hu, Yining and Bai, Yue and Li, Xiang and Yao, Yuan},
  booktitle = {European Conference on Computer Vision (ECCV)},
  year      = {2024}
}

@inproceedings{zhang2016icip,
  author    = {Zhang, Lei and Yang, Fan and Zhang, Y. D. and Zhu, Y. J.},
  title     = {Road Crack Detection Using Deep Convolutional Neural Network},
  booktitle = {Proceedings of the IEEE International Conference on Image Processing (ICIP)},
  pages     = {3708--3712},
  year      = {2016}
}

@article{zou2018deepcrack,
  author  = {Zou, Qin and Zhang, Zheng and Li, Qingquan and Qi, Xin and Wang, Qian and Wang, Song},
  title   = {DeepCrack: Learning Hierarchical Convolutional Features for Crack Detection},
  journal = {IEEE Transactions on Image Processing},
  volume  = {28},
  number  = {3},
  pages   = {1498--1512},
  year    = {2019},
  doi     = {10.1109/TIP.2018.2878966}
}

@article{zou2012cracktree,
  author  = {Zou, Qin and Cao, Yuchen and Li, Qingquan and Mao, Qingzhou and Wang, Song},
  title   = {CrackTree: Automatic Crack Detection from Pavement Images},
  journal = {Pattern Recognition Letters},
  volume  = {33},
  number  = {3},
  pages   = {227--238},
  year    = {2012},
  doi     = {10.1016/j.patrec.2011.11.004}
}

@article{staal2004drive,
  author  = {Staal, Joes and Abr{\`a}moff, Michael D. and Niemeijer, Meindert and Viergever, Max A. and van Ginneken, Bram},
  title   = {Ridge-based Vessel Segmentation in Color Images of the Retina},
  journal = {IEEE Transactions on Medical Imaging},
  volume  = {23},
  number  = {4},
  pages   = {501--509},
  year    = {2004},
  doi     = {10.1109/TMI.2004.825627}
}

@inproceedings{ma2021selfsupervised,
  author    = {Ma, Yuxin and Hua, Yang and Deng, Hanming and Song, Tao and Wang, Hao and Xue, Zhengui and Cao, Heng and Ma, Ruhui and Guan, Haibing},
  title     = {Self-Supervised Vessel Segmentation via Adversarial Learning},
  booktitle = {Proceedings of the IEEE/CVF International Conference on Computer Vision (ICCV)},
  pages     = {7536--7545},
  year      = {2021}
}

@article{flowsdf,
  author  = {Bogensperger, Lea and Narnhofer, Dominik and Falk, Alexander and Schindler, Konrad and Pock, Thomas},
  title   = {FlowSDF: Flow Matching for Medical Image Segmentation Using Distance Transforms},
  journal = {International Journal of Computer Vision},
  volume  = {133},
  number  = {7},
  pages   = {4864--4876},
  year    = {2025}
}

@inproceedings{kingma2015adam,
  author    = {Kingma, Diederik P. and Ba, Jimmy},
  title     = {Adam: A Method for Stochastic Optimization},
  booktitle = {International Conference on Learning Representations (ICLR)},
  year      = {2015}
}

@inproceedings{mou2024t2iadapter,
  author    = {Mou, Chong and Wang, Xin and Xie, Lingxi and Wu, Yaowei and Zhang, Jie and Qi, Zibin and Shan, Ying},
  title     = {T2I-Adapter: Learning Adapters to Dig Out More Controllable Ability for Text-to-Image Diffusion Models},
  booktitle = {Proceedings of the AAAI Conference on Artificial Intelligence},
  year      = {2024},
  volume    = {38},
  pages     = {4296--4304}
}

@inproceedings{lei2024enriching,
  author    = {Lei, Qi and Zhong, Jiajia and Dai, Qionghai},
  title     = {Enriching Information and Preserving Semantic Consistency in Expanding Curvilinear Object Segmentation Datasets},
  booktitle = {Proceedings of the European Conference on Computer Vision (ECCV)},
  year      = {2024},
  pages     = {233--250},
  publisher = {Springer}
}

@inproceedings{stucki2023topologically,
  title={Topologically faithful image segmentation via induced matching of persistence barcodes},
  author={Stucki, Nico and Paetzold, Johannes C and Shit, Suprosanna and Menze, Bjoern and Bauer, Ulrich},
  booktitle={International Conference on Machine Learning},
  pages={32698--32727},
  year={2023},
  organization={PMLR}
}
\end{document}